\documentclass[11pt]{article}

\usepackage[final]{acl}

\usepackage{times}
\usepackage{latexsym}
\usepackage{booktabs}
\usepackage[T1]{fontenc}

\usepackage[utf8]{inputenc}

\usepackage{microtype}
\usepackage{inconsolata}

\usepackage{graphicx}
\usepackage{multirow}
\usepackage{amsmath}
\usepackage{amssymb}
\newcommand{\mtok}{m_{\mathrm{tok}}}
\title{Gated Tree Cross-Attention for Checkpoint-Compatible Syntax Injection in Decoder-Only LLMs}

\author{Xinyu Gao\textsuperscript{1},  Shaonan Wang\textsuperscript{2},  Nai Ding\textsuperscript{1\thanks{Corresponding author: Nai Ding}} \\
\textsuperscript{1}College of Biomedical Engineering \& Instrument Science, Zhejiang University \\
\textsuperscript{2}Department of Language Science and Technology, The Hong Kong Polytechnic University \\
\texttt{\{xinyu\_gao, ding\_nai\}@zju.edu.cn} \\
\texttt{shaonan.wang@polyu.edu.hk}
}

\begin{document}
\maketitle
\begin{abstract}
Decoder-only large language models achieve strong broad performance but are brittle to minor grammatical perturbations, undermining reliability for downstream reasoning. However, directly injecting explicit syntactic structure into an existing checkpoint can interfere with its pretrained competence. We introduce a checkpoint-compatible gated tree cross-attention (GTCA) branch that reads precomputed constituency chunk memory while leaving backbone architecture unchanged. Our design uses a token update mask and staged training to control the scope and timing of structural updates. Across benchmarks and Transformer backbones, GTCA strengthens syntactic robustness beyond continued training baselines without compromising Multiple-Choice QA performance or commonsense reasoning, providing a practical checkpoint-compatible route to more syntax-robust decoder-only LLMs\footnote{\url{https://github.com/Pineandgrass/GatedTreeCrossAttention}.}. 
\end{abstract}

\section{Introduction}

Transformer-based pre-trained language models (PLMs) now underpin most Natural Language Understanding (NLU) and Natural Language Generation (NLG) systems because scaled self-attention transfers broadly across tasks \citep{vaswani-2017-attention1,devlin-2019-bert2,Brown-2020-language5,liu-2019-roberta3,Raffel-2020-exploring4,touvron-2023-llama6}. However, for reliable deployment, it is not enough to score well on aggregate benchmarks: a model should keep its decision stable under small grammatical perturbations and syntactic alternations that preserve meaning. In practice, decoder-only LLMs can flip preferences on minimally different inputs (e.g., agreement or licensing minimal pairs), a failure mode that users experience as “same meaning, different answer” and that can cascade into downstream reasoning errors \cite{ribeiro-etal-2020-beyond}.

This mismatch is visible on the targeted stress tests. Natural Language Inference (NLI) models trained in Multi-Genre Natural Language Inference (MNLI) can appear strong while relying on shallow heuristics, and fail in Heuristic Analysis for NLI Systems (HANS) where correct labels require syntactic generalization \citep{gururangan-2018-annotation41,mccoy-2019-right42}. Similarly, the Benchmark of Linguistic Minimal Pairs (BLiMP) reveals persistent gaps in structure-sensitive constraints (e.g., islands), with GPT2 far below human performance \citep{warstadt-2020-blimp-benchmark16}.

Why does this brittleness persist even as the models seem linguistically informed? The probing work shows that the non-trivial syntactic structure is recoverable from hidden states \citep{hewitt-2019-structural7} and is reflected in layers and attention patterns \citep{clark-2019-bert8,eisape-2022-probing14,muller-2022-spectral13}. However, encoding is not usage. Recoverability does not ensure that a model uses structure in its predictions, and probes can be misleading under redundancy or artifacts \citep{die-2025-probingsyntaxlargelanguage19,agarwal-2025-mechanisms18,tucker-2022-syntax17}. From an adaptation standpoint, naive structure injection can also interfere with pretrained representations, triggering instability and catastrophic forgetting during continued training \citep{iwamoto-2023-incorporating22}. This motivates controllable syntax injection, we want to expose structure in a way that the model can selectively exploit, without rewriting the checkpoint.

Our solution is a minimally invasive side path: a checkpoint-compatible gated cross-attention (GTCA) branch that lets token states read cached constituency chunk memory while leaving the backbone architecture untouched. The gate learns when to trust the structural signal, turning the structure into a regulated update rather than a hard constraint. We further restrict interference with token update masking and a three-stage training schedule.

Evaluation covers multiple-choice QA (MCQA) benchmarks, including CLOTH \citep{xie-2018-large28} and MMLU \citep{hendrycks-2021-measuringmassivemultitasklanguage27}, syntactic benchmarks, including BLiMP \citep{warstadt-2020-blimp-benchmark16} and CoLA \cite{cola} from GLUE \citep{wang-etal-2018-glue}, and commonsense reasoning tasks, including HellaSwag \citep{zellers-etal-2019-hellaswag} and WinoGrande \citep{WinoGrande}. GTCA consistently improves syntax-focused results while maintaining MCQA accuracy and overall capability across different backbones. Layer-wise probe of Unlabeled Undirected Attachment Score (UUAS) further connects these gains to a more syntax-consistent internal structure \citep{hewitt-2019-structural7,muller-2022-spectral13,limisiewicz-2021-introducingorthogonalconstraintstructural11}. Specifically, GTCA increases the accuracy of BLiMP from 78.58 to 83.12 in Qwen-2.5-7B and from 79.95 to 84.61 in Llama-3-8B, despite leaving the backbone architecture unchanged.

Our major contributions are as follows.

\textbullet\ We propose GTCA, a checkpoint-compatible gated cross-attention side branch that injects hierarchical constituency information via cached chunk memory without modifying backbone architecture.

\textbullet\ We introduce two stabilization mechanisms, token update masking and staged training, to control interference during continued training.

\textbullet\ Across two decoder-only backbones with comparable capacities, GTCA yields consistent syntax improvements while preserving or improving broad competence.

\section{Related Work}
\subsection{Injecting Explicit Syntactic Structure into Pre-Trained Transformers}

A long line of work has incorporated syntax as an explicit inductive bias in neural NLP, ranging from tree-structured composition to attention-based structure integration. Early architectures encoded hierarchical structure through tree-structured recurrence \citep{tai-2015-improved29}. Later approaches injected syntactic information into attention-based models by biasing attention towards syntactic relations, such as biasing self-attention using syntactic relations \citep{strubell-2018-linguistically30} or designing dependency-aware attention mechanisms for sequence-to-sequence modeling \citep{bugliarello-2020-enhancing31}. With the rise of PLMs, researchers have increasingly asked whether explicit trees still help when models already capture syntax implicitly, and how to leverage trees without expensive retraining. Representative efforts have included plug-in-style syntax integration into pre-trained Transformers \citep{bai-2021-syntax32} and systematic studies on when syntactic structure is beneficial under different injection strategies and tasks \citep{sachan-2021-syntax33}. Taken together, previous work spans architectural biases, objective-level supervision, and structure-native pretraining.

A related line of work injected syntax through training objectives rather than architectural rewrites, e.g., syntax-guided contrastive learning \citep{zhang-2022-syntax34}, or continuing training pipelines that explicitly considered optimization stability and catastrophic forgetting when incorporating syntactic knowledge \citep{iwamoto-2023-incorporating22}. In parallel, modeling native language structure on a scale aims to induce or enforce hierarchical structure during pretraining \citep{hu-2024-generative35,zhao-2024-dependency36}, and graph-based formulations offer a broader perspective of structured modeling \citep{plenz-2024-graph37}. At the same time, cautionary evidence suggests that tree bias alone is not sufficient. Its effectiveness depends on how the structure is represented and integrated \citep{ginn-2024-treetransformersineffectivemodel20}. These findings motivate checkpoint-compatible, minimally invasive structure injection for decoder-only LLMs, where hierarchical signals are exposed through an auxiliary pathway and integrated in a way that the pre-trained backbone can reliably exploit.

\subsection{Probing Syntactic Representations and Training-Efficient Adaptation}

Probing has been central to characterizing syntactic information in neural language models and to clarifying the mechanisms-outcomes tension. Structural probing showed that dependency-like geometry could be recovered from contextual embeddings \citep{hewitt-2019-structural7}, while attention and layer-wise analyses have revealed systematic associations between model components and linguistic relations or processing stages \citep{jawahar-2019-bert10,tenney-2019-bertrediscoversclassicalnlp9,clark-2019-bert8}. Subsequent work improved probe formulations and constraints \citep{limisiewicz-2021-introducingorthogonalconstraintstructural11,muller-2022-spectral13}, extended probing to incremental parse states in autoregressive settings \citep{eisape-2022-probing14}, and proposed richer representational codes such as polar probes \citep{die-2024-polarcoordinaterepresentssyntax15}. 

However, recent evidence has emphasized that representational recoverability does not guarantee behavioral usage. Redundancy and probe artifacts can produce misleading conclusions, motivating intervention-style and dropout-based analyses \citep{tucker-2022-syntax17,agarwal-2025-mechanisms18,die-2025-probingsyntaxlargelanguage19}, while other work argued for a richer generative-style structure than previous probes captured \citep{kennedy-2025-evidence21}. In parallel, practical checkpoint adaptation has been shaped by parameter-efficient fine-tuning methods such as LoRA \citep{hu-2021l-oralowrankadaptationlarge25} and QLoRA \citep{Dettmers-2023-qloraefficientfinetuning26}, prompt-based alternatives including prefix and prompt tuning \citep{li-2021-prefix38,lester-2021-power39,liu-2022-p40}, and stability-oriented components such as gated attention variants that improve optimization dynamics \citep{qiu-2025-gatedattentionlargelanguage24}. Together, this literature motivates approaches that control interference during continued training and that evaluate syntax gains through both targeted behavioral tests and layer-wise structural probes. These principles are reflected in our staged training pipeline and our joint evaluation. 
\begin{figure}[t] 
    \includegraphics[width=\columnwidth]{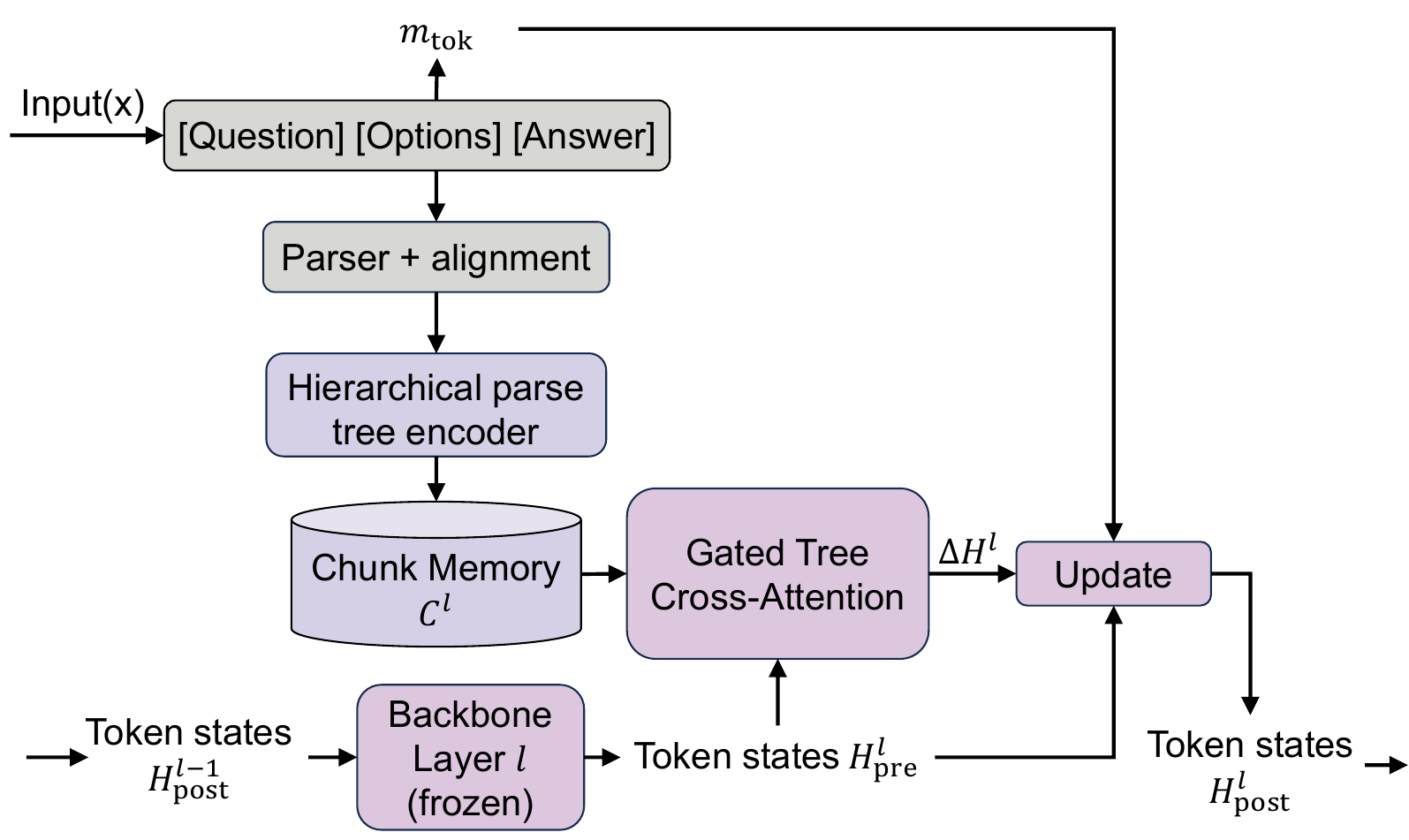} 
\caption{Overview of GTCA (Gated Tree Cross-Attention) structural injection.Given an input (question, options, and answer field), we run an offline parser with span alignment and encode the resulting hierarchy with a parse-tree encoder to build per-layer chunk memory $C^\ell$. At Transformer layer $\ell$, the pre-update token states $H_{\mathrm{pre}}^\ell$ query $C^\ell$ through gated tree cross-attention to produce a structural residual update $\Delta H^\ell$. This update is then added to obtain the post-update token states $H_{\mathrm{post}}^\ell$, which are passed to the next layer.
}
    \label{fig:1}
\end{figure}

\begin{figure*}[t] 
    \centering
    \includegraphics[width=0.8\textwidth]{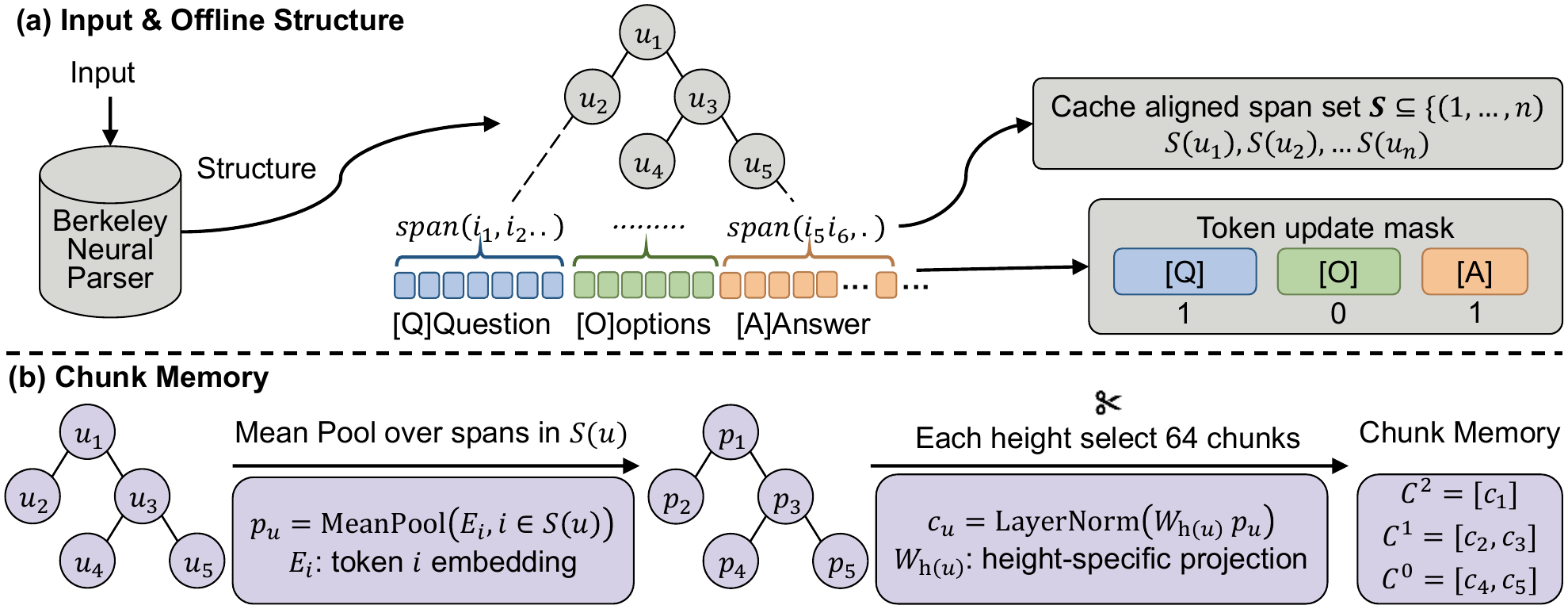} 
\caption{A constituency parser produces cached span-aligned nodes $S(u)$ and update mask. Chunk memory is computed by mean-pooling over $S(u)$ followed by a height-specific projection and LayerNorm.}
    \label{fig:2}
\end{figure*}

\section{Method}
We augment a decoder-only Transformer with an auxiliary, checkpoint-compatible pathway that injects hierarchical syntactic structure through gated cross-attention. The backbone architecture remains unchanged, and the structural pathway is attached as a checkpoint-compatible forward wrapper and can be enabled or disabled without re-training from scratch (Figure \ref{fig:1}).
\subsection{Precomputed Parse Tree Structure Aligned to Token Spans}
\label{Sec3.1}
\paragraph{Parse tree memory.} As shown in Figure \ref{fig:2} (a), given a tokenized input sequence $x_{1:n}$, we precompute a constituency parse tree offline, using Berkeley Neural Parser \cite{kitaev-klein-2018-constituency}. Each tree node (chunk) $u$ is aligned to a token span $S(u) \subseteq \{1,\dots,n\}$ and stores its chunk type and parent or child relations. When a parser word is split into multiple Byte Pair Encoding (BPE) tokens, we treat its subword block as a subnode \citep{BPE}. For efficiency and determinism, we cache the parsed structure and span alignments, indexed by a hash of the input token IDs. This design eliminates parsing overhead during training and ensures that the same input always retrieves the same tree.

\paragraph{MCQA-specific structure and token update mask.} Our training uses a MCQA format, where the input contains a question region and a set of candidate options (Figure \ref{fig:2} (a)). To prevent the structural branch from altering the hidden states of candidate options (which would change their relative likelihoods under likelihood-based scoring), we treat option tokens as read-only memory under the structural pathway. 

Concretely, we define a binary token update mask $\mtok \in \{0,1\}^n$ that enables structural updates only for question tokens and the answer field, while forcing $\mtok=0$ for option tokens. This prevents option preference drift during continued training and confines structural intervention to the parts of the input where syntactic reasoning is intended to operate.

\subsection{Hierarchical Parse Tree Encoder for Chunk Memory Construction}
\label{Sec3.2}
For the token $i$, we denote the token embedding by $E^i \in \mathbb{R}^{1 \times d}$, where $d$ is the hidden size. The precomputed constituency tree contains a set $U$ of chunks, and each chunk $u \in U$ is aligned to a token span $S(u) \subseteq \{1,\dots,n\}$. And we define the height of the chunk $u$ as  $h(u) =D-depth(u)$, where $depth(u)$ is the number of edges from the root to $u$ and $D$ is the maximum depth over the leaf tokens in the constituency tree for the current input. Thus, the deepest leaf tokens have height 0 and the root has the maximum height (Figure \ref{fig:2} (b)).
For each chunk $u$ at height $h(u)$, we first aggregate token states over its span through mean pooling:
\begin{equation}
  \label{eq:1}
p_u = \text{MeanPool}\left(\overline{E_i}, i \in S(u)\right)
\end{equation}
We then apply a height specific linear projection based on tree height and normalize the resulting chunk embeddings for stability:
\begin{equation}
  \label{eq:2}
c_u = \text{LayerNorm}\left(W_{h(u)} \, p_u\right)
\end{equation}
where $W_{h(u)} \in \mathbb{R}^{d \times d}$ is shared by chunks of height $h(u)$. To couple tree height with Transformer layers, we use the layer index $\ell$ to select a height level for chunk memory, then the per-chunk embeddings are stacked to form the chunk memory matrix at layer $\ell$:
\begin{equation}
  \label{eq:n}
\tilde{\ell} = \min(\ell, D), \quad 
C^\ell = \left[ c_u \right]_{u \in U, \, h(u) = \tilde{\ell}}
\end{equation}

Because the maximum chunk height $D$ varies by input, when $D < \ell$ we reuse the top-level chunk memory for higher Transformer layers. To keep computation and memory manageable, we retain at most $K = 64$ chunks per height. Concretely, we traverse the constituency tree in a left-to-right breadth-first order, and for each height $h$ we keep the first $K$ chunks encountered.

This layer-aligned design, where the chunk $C^\ell$ is consumed by the Transformer layer $\ell$, is intended to encourage an intuitive hierarchy of structural utilization, such that the higher layers draw on the higher-level parse chunks while the lower layers primarily rely on more local structure. We empirically probe this effect via UUAS (Section \ref{Sec 4.3}). Controlled ablations on Qwen-2.5-7B for the projection scheme, span pooling, and memory size are reported in Appendix \ref{appendixAdditional}.

\subsection{Gated Parse Tree Cross-Attention Injected via Wrappers}
We inject parse tree structure through an auxiliary cross-attention branch implemented as a forward wrapper, using head-wise gated attention \citep{qiu-2025-gatedattentionlargelanguage24}. The headwise gate is adopted because previous work showed that it introduces far fewer parameters than the elementwise gate while achieving comparable perplexity and downstream accuracy, thus reducing memory overhead and stabilizing training.

At Transformer layer $\ell$, let $C^\ell$ denote the chunk memory at height $\ell$ (defined in Section \ref{Sec3.2}), and $\overline{H^\ell_{\mathrm{pre}}}$ denote the pre-normalization token hidden states before the GTCA update. We use projection matrices $W_Q^\ell, W_K^\ell, W_V^\ell$ and $W_G^\ell$, then compute queries, keys, values, gate score, and attention output with head dimension $d_h$:
\begin{equation}
  \label{eq:3}
    \begin{split}
    Q^\ell &= \overline{H^\ell_{\mathrm{pre}}} W_Q^\ell, \quad
    K^\ell = C^\ell W_K^\ell, \\
    V^\ell &= C^\ell W_V^\ell, \quad
    G^\ell = \overline{H^\ell_{\mathrm{pre}}} W_G^\ell,
    \end{split}
\end{equation}
\begin{equation}
  \label{eq:4}
\text{Attn}^\ell = \text{softmax}\left(\frac{Q^\ell (K^\ell)^T}{\sqrt{d_h}} + M_{\text{causal}}\right) V^\ell
\end{equation}
where $M_{\text{causal}}$  masks out chunks whose right boundary exceeds the current token position, ensuring that the cross-attention never accesses future information. Following prior work on head-wise gated attention, $G^\ell$ serves as the gating logit after the attention output:
\begin{equation}
  \label{eq:5}
\text{Gated\_Attn}^\ell = \text{Attn}^\ell \odot \sigma(G^\ell)
\end{equation}
where $\sigma$ is the logistic sigmoid and $\odot$ denotes elementwise multiplication. Here, $G^\ell$ provides one logit per attention head, broadcast over $d_h$. The gated heads are then merged and projected, and we use them to update token hidden states:
\begin{equation}
  \label{eq:6}
\Delta H^\ell = \text{Merge}(\text{Gated\_Attn}^\ell) W_O^{ca,\ell}
\end{equation}
where $W_O^{ca,\ell}$ is the cross-attention output projection that maps the merged multi-head output back to the model hidden size, and token hidden states are updated as:
\begin{equation}
  \label{eq:7}
H^\ell_{\mathrm{post}} \leftarrow H^\ell_{\mathrm{pre}} + \alpha_{\text{struct}} \bigl(\mtok[:,\text{None}] \odot \Delta H^\ell\bigr)
\end{equation}
where the token update mask $\mtok \in \{0,1\}^n$ is defined in Section \ref{Sec3.1} and the structural coefficient $\alpha_{\text{struct}}$ follows the schedule in Section \ref{Sec 3.4}. Here $\mtok[:,\mathrm{None}]$ denotes broadcasting along the hidden dimension, so that each row of $\Delta H^\ell$ is multiplied by the corresponding mask entry before the masked update is added to $H_{\mathrm{pre}}^\ell$ to obtain $H_{\mathrm{post}}^\ell$.

\subsection{Three-Stage Training to Balance Syntax Gains and Retention}
\label{Sec 3.4}
We use a staged training scheme to balance syntax specialization and the retention of broad general ability.

Stage 1: Adaptation of the task. We apply LoRA adapters to the backbone projection matrices to fit the MCQA-style training objective and the corresponding instruction format with limited trainable parameters. During this stage, the structural pathway is disabled by setting $\alpha_{\text{struct}}$=0, so the adaptation does not rely on the injected structure.

Stage 2: Integration of the structure. We freeze the backbone and LoRA adapters, train only the parse tree encoder, cross-attention, and gating modules. To avoid abrupt distribution shifts, $\alpha_{\text{struct}}$ is warmed up from 0 to $\alpha_{\text{struct}}^*$ over the first 10\% of Stage 2 steps. 

Stage 3: Joint refinement. We jointly train the structural modules and LoRA adapters while keeping the rest of the backbone frozen. This stage reconciles structure-sensitive updates with the task-adapted representations, and aims to preserve the syntax gains from Stage 2 without incurring expensive full fine-tuning.

Overall, GTCA's staged design enables efficient structure-aware adaptation by introducing only an $O(nK)$ cross-attention overhead, which remains cheaper than the $O(n^2)$ self-attention in the backbone, while keeping the pre-trained model largely intact and the process inspectable.

\section{Experiments}
\subsection{Experimental Setup}
\paragraph{Model variants.} We evaluated four checkpoint-based variants instantiated from a common pretrained backbone: (i) Backbone, the original pretrained model without continued training; (ii) LoRA-only, low-rank adapter task adaptation while disabling the structural pathway ($\alpha_{\text{struct}}=0$). This baseline is trained for the full number of optimization steps and therefore is not the intermediate Stage1 checkpoint; (iii) Direct-Joint, single-stage training of structural modules and LoRA adapters in Stage 3 setup, skipping Stages 1 to 2; and (iv) GTCA, our proposed gated tree cross-attention model trained with the three-stage pipeline (Section \ref{Sec 3.4}).

Unless otherwise stated, all experiments in the main manuscript are reported in both Qwen-2.5-7B \citep{qwen2025qwen25technicalreport} and Llama-3-8B \citep{grattafiori2024llama3herdmodels}, using the same maximum sequence length and prompts, computing budgets, and the number of training steps, while keeping the default tokenizer of each backbone. Additionally, Appendix \ref{appendix1} provides parameter counts for the 8 variants (4 variants × 2 backbones), along with dataset details, the overall compute budget and computing infrastructure.

\begin{table}[t]
\centering
\resizebox{0.8\columnwidth}{!}{
\begin{tabular}{lllllll}
\specialrule{1.2pt}{0pt}{0pt}
           & \#Train       & \#Dev      & \#Test         & \#Class    \\
\midrule
CLOTH      & 76850         & 11067      & 11516          & 4      \\
MMLU       & 99842         & 285        & 14042          & 4          \\
CoLA       & 8551          & 527        & 516            & 2          \\
BLiMP      & /             & /          & 67000          & /          \\
HellaSwag  & 39905         & 10042      & 10003          & 4          \\
WinoGrande & 40398         & 1267       & 1767           & 2          \\
\specialrule{1.2pt}{0pt}{0pt}
\end{tabular}
}
  \caption{The statistics and segmentation usage of the dataset, only the first two datasets were used for training.}
  \label{tab:statistics}
\end{table}

\paragraph{Datasets and preprocessing.} Our continued training used CLOTH and MMLU in a unified MCQA format \citep{hendrycks-2021-measuringmassivemultitasklanguage27,xie-2018-large28}. We evaluated on CLOTH, MMLU, BLiMP, CoLA, HellaSwag, and WinoGrande. MMLU and CLOTH are included to assess MCQA performance. BLiMP and CoLA primarily test targeted syntactic competence. HellaSwag and WinoGrande primarily measure more general-purpose capabilities, such as commonsense and broader reasoning. Dataset split statistics and usage are summarized in Table \ref{tab:statistics}. 

For structure-aware variants, we precomputed constituency parse trees offline, cached the resulting parse tree memory. To ensure that all examples could be successfully parsed, we applied a minimal text normalization to the raw data by collapsing consecutive whitespace into a single space, and modeled each sentence independently in all the experiments. For MCQA prompts, we parsed each field independently (question, each option, and the answer prefix), rather than parsing the full concatenated prompt. We then concatenated the resulting span-aligned chunks in the original prompt order. In addition, Appendix \ref{appendixB} provides the prompts used for each benchmark along with the corresponding dataset, checkpoint licensing and terms of use.

\begin{table}[t]
\resizebox{\columnwidth}{!}{
\begin{tabular}{lllll}
\specialrule{1.2pt}{0pt}{0pt}
Stage  & Learning rate & LoRA rank/$\alpha_{\text{lora}}$ & Max\_length & $\alpha_{\text{struct}}$ \\
  \\
\hline
Stage1 & 5e-5          & 16/32                & 1024       & 0  \\
Stage2 & 3e-5          & 0/0                  & 1024       & 0 to 0.15 \\
Stage3 & 5e-5          & 16/32                & 1024       & 0.15        \\

\specialrule{1.2pt}{0pt}{0pt}
\end{tabular}
}
  \caption{Hyperparameters at different stages.}
  \label{tab:Hyperparameters}
\end{table}

\paragraph{Training details.} We followed the three-stage training scheme in Section \ref{Sec 3.4}. The key hyperparameters are reported in Table \ref{tab:Hyperparameters}. We selected the structural coefficient $\alpha_{\text{struct}}^*=0.15$ through grid search (Appendix \ref{appendixD}).

Due to the checkpoint-compatible design, the absolute differences between the backbone parameters before and after training are 0.0 for both backbones, indicating that the original backbone checkpoint parameters remain unchanged.

\paragraph{Evaluation details.} For MCQA-style benchmarks, given a prompt $prompt$ and $r$ candidate options $\{option^r\}$, each option is scored by its conditional log-likelihood under the model. More precisely, it is computed as the sum of token-level log-probabilities when generating the option conditioned on the prompt and previously generated tokens in the option:
\begin{equation}
\begin{aligned}
\label{eqq}
&\mathrm{score}\!\left(option^j \mid prompt\right)= \\
&\sum_{i=1}^{|option^j|}
\log p_\theta\!\left(option_i^j \mid prompt,\right. 
\left.option_{<i}^j\right)
\end{aligned}
\end{equation}
where $j\in\{1,\ldots,r\}$ indexes candidate options and $i$ indexes tokens within $option^{j}$. And our token update mask can keep the option log-likelihoods in Equation \ref{eqq} comparable across candidates, thereby cleanly reflecting the influence of syntactic structure in the tree. The prediction was the option with the highest score:
\begin{equation}
\hat{j} = \arg\max_{j \in \{1,\dots,r\}}
\mathrm{score}\!\left(option^{j} \mid prompt\right)
\end{equation}
When option lengths vary strongly, length normalized option scoring is preferable, but in our setting it leaves the method ranking unchanged.

Under $k$-shot in-context evaluation, we prepended $k$ demonstration examples to the prompt at inference time, and applied the same scoring rule. For BLiMP, we instead used the pairwise likelihood preference, considering an item as correct if $\log p(s^+) > \log p(s^-)$, where $s^{+}$ is the good sentence and $s^{-}$ is the bad sentence. Following standard practice, the Matthews correlation coefficient (MCC, \%) is used for CoLA and accuracy (\%) for all other benchmarks. And for all our experiments in this paper, we reported mean scores over 5 runs.  

\begin{table}[t]
\centering
\resizebox{\columnwidth}{!}{%
\begin{tabular}{l cc cc cc}
\specialrule{1.2pt}{0pt}{0pt}
\multirow{2}{*}{\centering Model} & \multicolumn{2}{c}{MCQA} & \multicolumn{2}{c}{Syntax} & \multicolumn{2}{c}{Commonsense} \\

\cmidrule(lr){2-3}\cmidrule(lr){4-5}\cmidrule(lr){6-7}
      & CLOTH & MMLU & BLiMP & CoLA & HellaSwag & WinoGrande \\
      & (0-shot) & (0-shot) &  & (0-shot) & (10-shot) & (5-shot) \\
\hline

\multicolumn{7}{c}{\textbf{Backbone: Qwen-2.5-7B}} \\
\hline
Backbone     & 74.60 & 69.98 & 78.58 & \textbf{ 59.71}$^\dagger$ & 61.79 & 73.01 \\
LoRA-only    & 83.71 & 70.12 & 80.87 & 53.31          & 62.74 & \textbf{75.10} \\
Direct-Joint & \textbf{ 84.80}$^\dagger$ & 69.55 & 80.27 & 52.75 & 62.04 & 74.23 \\
GTCA         & 83.98 & \textbf{ 71.02}$^\dagger$ & \textbf{ 83.12}$^\dagger$ & 56.59 & \textbf{ 63.23}$^\dagger$ & 74.95 \\
\hline

\multicolumn{7}{c}{\textbf{Backbone: Llama-3-8B}} \\
\hline
Backbone     & 41.79 & 54.14 & 79.95 & 53.57 & 61.76 & 76.72 \\
LoRA-only    & 81.38 & 53.78 & 80.70 & 53.68 & 63.24 & 76.01 \\
Direct-Joint & 80.46 & 52.36 & 80.36 & 52.46 & 62.86 & 76.98 \\
GTCA         & \textbf{ 82.74}$^\dagger$ & \textbf{ 54.97}$^\dagger$ & \textbf{ 84.61}$^\dagger$ & \textbf{ 56.69}$^\dagger$ & \textbf{ 64.85}$^\dagger$ & \textbf{ 77.89}$^\dagger$ \\
\specialrule{1.2pt}{0pt}{0pt}
\end{tabular}%
}
  \caption{Total scores on CLOTH, BLiMP, MMLU, CoLA, HellaSwag, and WinoGrande for Backbone, LoRA-only, Direct-Joint, and GTCA. Boldface with $\dagger$ indicates the best result when the improvement is significant under paired tests with Holm--Bonferroni correction ($p < 0.05$ after correction).}
  \label{tab:main result}
\end{table}

\subsection{Multiple-Choice QA Performance on CLOTH}
The CLOTH and MMLU results in both Qwen-2.5-7B and Llama-3-8B are shown in Table \ref{tab:main result}. Notably, the Llama-3-8B Backbone starts much lower on CLOTH, suggesting it is less well aligned with our instruction-style MCQA prompt format before adaptation. LoRA-only substantially improves in-format accuracy over the Backbone method on both backbone models, confirming that parameter-efficient continued training effectively adapted the model to the MCQA format. 

Importantly, adding parse-tree memory via gated cross-attention preserves in-format performance across backbone models. Our method GTCA achieves 83.98/71.02 (CLOTH/MMLU) on Qwen-2.5-7B and 82.74/54.97 on Llama-3-8B, outperforming both the Backbone and the LoRA-only baseline, indicating that the token update mask and gated structural pathway do not impede optimization on training-format tasks. Direct-Joint attains slightly higher CLOTH accuracy on Qwen-2.5-7B, but, as shown next, exhibits a less favorable trade-off on retention and targeted syntactic generalization. 

\subsection{Retention, Robustness, and Syntactic Generalization}
The overall performance is reported in Table \ref{tab:main result} for both Qwen-2.5-7B and Llama-3-8B. On the syntax focused benchmarks BLiMP and CoLA, GTCA improves BLiMP accuracy over both Backbone and LoRA-only and partially recovers the CoLA degradation introduced by continued training across both backbones. In particular, the LoRA-only stage also improves BLiMP. This gain is attributed to the fact that some special task adaptation alone can already sharpen local grammatical preferences to some extent, even without explicit structural augmentation. 

In the general capability benchmarks HellaSwag and WinoGrande, GTCA delivers strong general capacity retention across both backbones. In both Qwen-2.5-7B and Llama-3-8B, GTCA achieves the best HellaSwag performance among the continued training baselines. For WinoGrande, GTCA matches LoRA-only in Qwen-2.5-7B and is strongest in Llama-3-8B. Although GTCA is not uniformly best on every general benchmark and backbone, it achieves top performance in most settings and remains competitive with the strongest baselines elsewhere. Together, these results indicate that staged training and gated integration produce measurable syntactic gains while maintaining broad general capability largely stable.

\subsection{Layer-Wise Syntactic Probing }
\label{Sec 4.3}
To connect behavioral improvements to representational changes, we performed layer-wise syntactic probing and evaluated it with the UUAS metric. For each model variant, we extracted hidden states from each Transformer layer and trained a lightweight structural probe to recover syntactic structure under controlled conditions. Probing was conducted on a random subset of 3,000 parse trees from the Penn Treebank \citep{ptb}. Following \citet{hewitt-2019-structural7}, we evaluated UUAS against dependency trees derived from Penn Treebank annotations, keeping the probing dataset, tree formalism, and probe configuration fixed across model variants to isolate representational differences.

\begin{figure}[t] 
    \centering
    \includegraphics[width=\columnwidth]{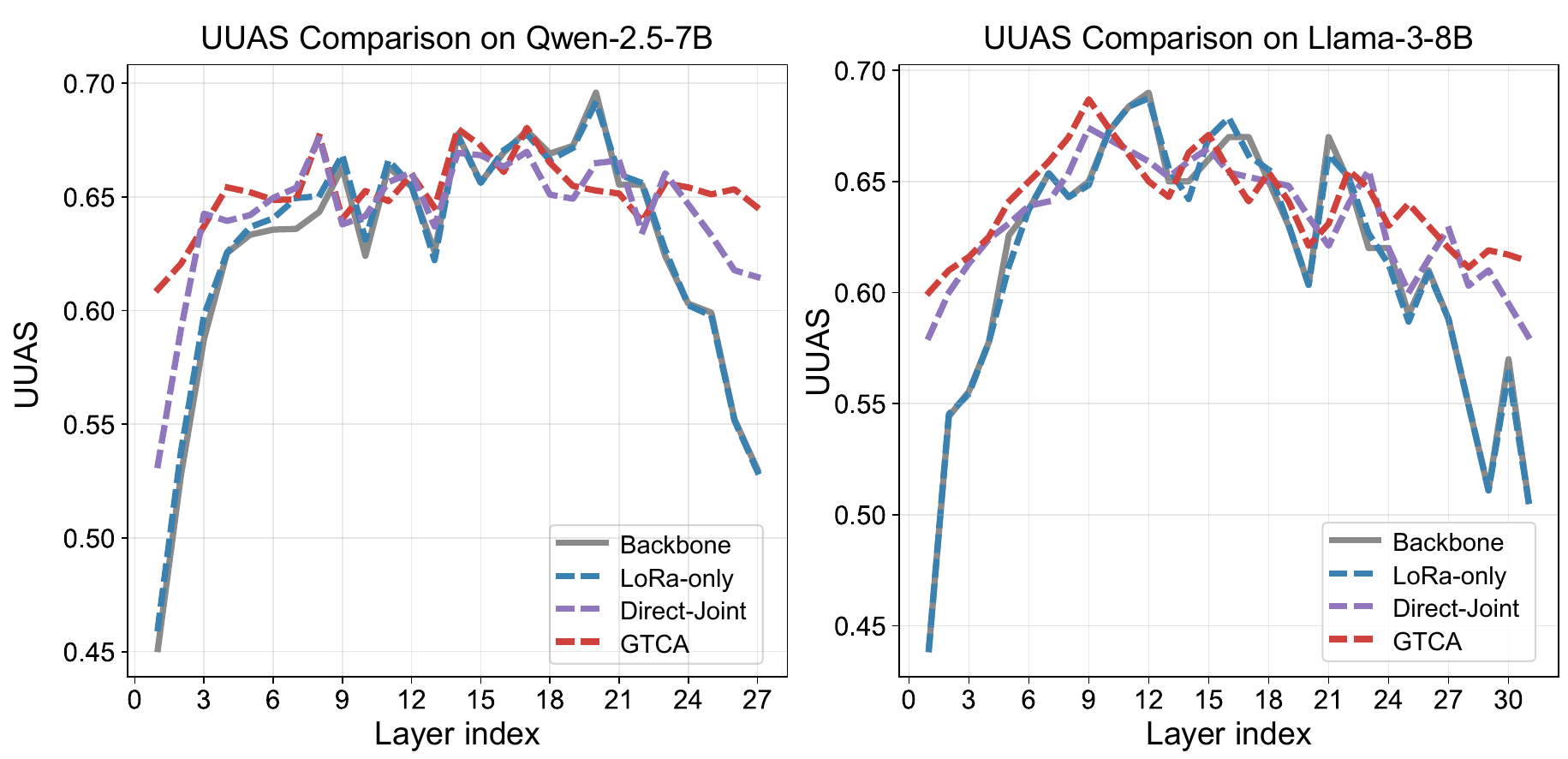} 
    \caption{UUAS of the Backbone, LoRA-only, Direct-Joint, and GTCA. UUAS is computed on intermediate layers, excluding the embedding layer.
}
    \label{fig:3}
\end{figure}

Figure \ref{fig:3} plots UUAS as a function of layer for Backbone, LoRA-only, Direct-Joint and GTCA in both Qwen-2.5-7B and Llama-3-8B. Across both backbones, structure injection increases syntactic recoverability primarily in the lower and upper layers. This layer-localized shift aligns with the BLiMP and CoLA improvements, providing evidence that GTCA changes internal geometry in a syntax-consistent direction rather than merely overfitting to the training objective.

\section{Analysis}
We analyzed which components drove the BLiMP and CoLA gains of syntax injection, and how they interacted with retention on general benchmarks. We first ablated gating, token update masking, and staged training, then tested whether improvements depended on coherent parse structure rather than generic auxiliary capacity. To isolate component effects without introducing confounding differences from backbone and tokenizer-specific differences, we conducted these follow-up analyses on Qwen-2.5-7B, while the main results already demonstrate consistent trends across both backbones.

\subsection{Ablations and Stage-Wise Effects}
Table \ref{tab:ablation} presents controlled ablations of the entire model. Overall, GTCA consistently delivers the best trade-off between syntactic generalization and retention, suggesting that each design choice contributes meaningfully rather than acting as redundant extra capacity. In particular, removing the gate degrades both the targeted syntax in BLiMP and the broad competence in MMLU, highlighting the role of the gate in controlling interference from the structural pathway. In contrast, disabling the token update mask leaves syntactic performance essentially unchanged but slightly weakens MCQA format stability and general retention. On a held-out set in Qwen-2.5-7B, the mean absolute change in pairwise option score gaps is 0.013 per token with the mask and 0.021 without it, consistent with the No Mask ablation in Table 4. This suggests that token update masking mainly stabilizes continued training under the MCQA interface rather than directly driving syntactic gains.

Table \ref{tab:stage} makes the stage-wise dynamics explicit and also includes a two stage pipeline that skips Stage 2. Stage 2 delivers the greatest improvement in BLiMP (about +2.6 points), showing that freezing the backbone and specializing structural modules is where syntactic generalization is most strongly acquired. This gain comes with a temporary drop in MMLU, but Stage 3 largely restores and even improves broad competence while preserving most of the syntactic improvement. This pattern suggests that Stage 2 and Stage 3 play complementary roles and that skipping either one yields a weaker tradeoff.

We further conducted a qualitative error analysis on BLiMP in Appendix \ref{appendixerror}, where the remaining errors are concentrated in filler gap dependencies and island effects, suggesting that the model can still over-rely on local surface compatibility and underweight global constraints that require tracking long-distance dependencies or respecting extraction islands.

\begin{table}[t]
\resizebox{\columnwidth}{!}{
  \begin{tabular}{lcccc}
    \specialrule{1.2pt}{0pt}{0pt}
    & \multicolumn{2}{c}{Multiple-Choice QA} & \multicolumn{2}{c}{Syntactic\ Robustness} \\
    \cline{2-3}\cline{4-5}
    Setting & CLOTH & MMLU & BLiMP & CoLA \\
    & (0-shot) & (0-shot) &  & (0-shot) \\
    \hline
    GTCA    & \textbf{ 83.98}$^\dagger$ & \textbf{ 71.02}$^\dagger$ & \textbf{ 83.12}$^\dagger$ & \textbf{ 56.59}$^\dagger$ \\
    No Gate & 82.84 & 69.80 & 81.68 & 55.04 \\
    No Mask & 83.71 & 70.50 & 83.12 & 55.68 \\
    \specialrule{1.2pt}{0pt}{0pt}
  \end{tabular}
  }
  \caption{Ablation results on key model components. No Mask sets $\mtok=1$ for all tokens. Boldface with $\dagger$ indicates the best result when the improvement is significant under paired tests with Holm--Bonferroni correction ($p < 0.05$ after correction).}
  \label{tab:ablation}
\end{table}
\begin{table}[t]
\resizebox{\columnwidth}{!}{
  \begin{tabular}{lcccc}
    \specialrule{1.2pt}{0pt}{0pt}
    & \multicolumn{2}{c}{Multiple-Choice QA} & \multicolumn{2}{c}{Syntactic Robustness} \\
    \cline{2-3}\cline{4-5}
    Stage & CLOTH & MMLU & BLiMP & CoLA \\
    & (0-shot) & (0-shot) &  & (0-shot) \\
    \hline
    Stage1 & 83.71 & 70.12 & 80.86 & 53.31 \\
    Stage2 & 83.70 & 68.07 & \textbf{ 83.50}$^\dagger$ & 54.13 \\
    Stage 1 then Stage 3 & 83.62 & 70.66 & 82.12 & 55.71 \\
    Stage3 & \textbf{ 83.98}$^\dagger$ & \textbf{ 71.02}$^\dagger$ & 83.12 & \textbf{ 56.59}$^\dagger$ \\
    
    \specialrule{1.2pt}{0pt}{0pt}
  \end{tabular}
  }
  \caption{Performance dynamics across specialization stages. Boldface with $\dagger$ indicates the best result when the improvement is significant under paired tests with Holm--Bonferroni correction ($p < 0.05$ after correction).}
  \label{tab:stage}
\end{table}

\begin{table}[t]
\resizebox{\columnwidth}{!}{
  \begin{tabular}{lcccc}
    \specialrule{1.2pt}{0pt}{0pt}
    & \multicolumn{2}{c}{Multiple-Choice QA} & \multicolumn{2}{c}{Syntactic Robustness} \\
    \cline{2-3}\cline{4-5}
    Tree& CLOTH & MMLU & BLiMP & CoLA \\
    & (0-shot) & (0-shot) &  & (0-shot) \\
\hline
    Weak parser& 80.66& 67.33& 78.25& 54.21\\
    Strong parser* & \textbf{ 83.98}$^\dagger$ & \textbf{ 71.02}$^\dagger$ & \textbf{ 83.12}$^\dagger$ & \textbf{ 56.59}$^\dagger$ \\
    Random trees & 70.52& 64.58& 74.36&50.36\\
    Permuted trees & 52.57& 65.55& 60.55& 52.66\\
    
    \specialrule{1.2pt}{0pt}{0pt}
  \end{tabular}
  }
  \caption{Tree quality and structure-faithfulness controls. Strong parser* corresponds to GTCA (full parser). Boldface with $\dagger$ indicates the best result when the improvement is significant under paired tests with Holm--Bonferroni correction ($p < 0.05$ after correction).}
  \label{tab:tree}
\end{table}

\subsection{Tree Quality and Structure-Faithfulness Controls}

Table \ref{tab:tree} tests whether performance depends on a linguistically meaningful structure and the Strong Parser column corresponds to GTCA. For the Weak Parser setting, we used the Berkeley parser variant with the smallest parameterization, while the remaining settings are obtained by degrading the strong parses by corrupting the tree structure (e.g., randomizing or permuting the resulting trees). As shown in the table, models benefit only when the parse trees are high quality and coherent. Replacing a strong parser with weaker ones consistently lowers both syntactic generalization and broad benchmark performance, indicating that low quality trees act more like noise than guidance. The corrupt structure further harms performance, with random trees causing a clear drop and permuted trees producing the most severe drop. These results support a structure-faithfulness interpretation that gains arise from consistent hierarchical signals in the parse tree memory rather than from an unconstrained auxiliary attention pathway. This interpretation is further supported by additional robustness analyses in Appendix \ref{appendixtree}, where GTCA remains more stable under controlled grammatical perturbations and mild surface noise than the continued training baselines.

\subsection{Interpreting the Structural Gate}
To summarize, a key design goal of GTCA is to expose syntactic information without forcing the backbone to always use it. The headwise gate provides an interpretable control signal for this purpose. For the token position $i$ in the layer $\ell$ and the attention head $head$, the effective structural contribution is modulated by
\begin{equation}
g^\ell_{head}(i) = \sigma(G^\ell_{head}(i))
\end{equation}

where $g^\ell_{head}(i)\in(0,1)$ can be interpreted as the confidence of the model in applying a structure-based update to this head. This makes GTCA behaviorally inspectable: when $g^\ell_{head}(i)$ is close to zero, the model effectively ignores parse-derived memory for that head; when it is close to one, the model performs a strong structure-conditioned update.

Importantly, this gating mechanism also clarifies why GTCA can improve syntactic benchmarks without consistently harming broad capability. Rather than injecting structure as a hard constraint, GTCA implements structure as a regulated residual. This matches the empirical role of the gate observed in ablations (Table \ref{tab:ablation}), removing the gate degrades both targeted syntactic generalization and retention, suggesting that the gate is not merely an extra capacity, but a stability mechanism that limits interference from potentially noisy or mismatched chunks.

\section{Conclusion}
This paper presents a checkpoint-compatible framework for strengthening syntactic competence in decoder-only LLMs by attaching a checkpoint-compatible gated tree cross-attention pathway (GTCA). The approach exposes hierarchical constituency structure through precomputed parse tree memory aligned to token spans, and modulates the cross-attention update with a learned headwise gate, while keeping the backbone architecture unchanged. To limit interference during continued training, the method combines token update masking in MCQA inputs with a three-stage training pipeline that separates task adaptation, structure specialization, and joint refinement. 
Empirically, GTCA improves syntactic generalization while preserving broad competence across multiple evaluations, and this trend holds across different backbones, including Qwen-2.5-7B and Llama-3-8B. Analysis further indicates that gating, token update masking and staged training are important for balancing syntactic gains and retention, and that performance is sensitive to tree quality and coherence, supporting a structure-faithfulness interpretation rather than a pure extra capacity effect.
\section*{Limitations}
The method relies on external constituency parsers and cached tree memories. As shown by the weak-parser and tree-corruption controls, low-quality or inconsistent parses can act as harmful noise, which makes the approach sensitive to parser accuracy and domain shift. This sensitivity can lead to brittle behavior and degraded outputs when the parse quality drops in real use. In addition, offline parsing and the auxiliary cross-attention branch introduce preprocessing, storage, and runtime overhead, which may become non-trivial for long-context settings or large-scale deployment and can increase latency and compute cost. 

Our evaluation focuses on English benchmarks and primarily measures broad competence using MCQA benchmarks. Moreover, extending GTCA to NLI-style training and stress-test setups (e.g., HANS-style controlled evaluations) is left for future work. And these results do not fully characterize downstream impacts on open-ended generation, multi-turn interaction, or long-context reasoning, where failure modes and reliability concerns may differ.

Finally, the current stability mechanisms are tailored to MCQA-style continued training, and broader validation across training paradigms, domains, and model families is required to establish robustness and generality, and to better understand how the method behaves under shifts in data, objectives, and model scale.
\section*{Ethics Statement}
We use only publicly available pretrained checkpoints and benchmark datasets and follow their licenses and terms of use. No new human participants or manual annotators were involved. 
\section*{Acknowledgment}
The authors would like to thank the anonymous reviewers for their helpful suggestions and comments. This work was supported by the Brain Science and Brain-like Intelligence Technology - National Science and Technology Major Project 2021ZD0204100 (2021ZD0204105 to N. D.).
\bibliography{main}
\appendix
\renewcommand{\thetable}{\arabic{table}}
\setcounter{table}{0}
\renewcommand{\tablename}{Appendix Table}

\section{Additional Ablations}
\label{appendixAdditional}
We ran a targeted ablation set on the tree encoder and memory construction. Then varied one factor at a time, and keep the rest identical to GTCA. The metrics are in Appendix Table \ref{tab:appendixAdditional}. All results are on Qwen-2.5-7B, and we set temperature=0 and disabled sampling (do\_sample=False).

The results show that a shared projection across heights reduces all metrics, this supports the idea that different granularities need different parameterization. Learned attention pooling gives a small gain on CoLA, K equals 128 chunks per height gives a small gain on BLiMP, but it adds parameters or introduces another learned attention module. Considering the joint trade-off between resource consumption and throughput we use GTCA default as in the main text.

\begin{table}[h]
\centering
\resizebox{\columnwidth}{!}{%
\begin{tabular}{lcccc}
\specialrule{1.2pt}{0pt}{0pt}
Setting & CLOTH & MMLU & BLiMP & CoLA \\
\specialrule{0.8pt}{0pt}{0pt}
GTCA default & \textbf{83.98} & \textbf{71.02} & 83.12 & 56.59 \\
Single shared projection across heights & 81.70 & 70.76 & 82.08 & \textbf{56.90} \\
Span pooling learned attention pooling & 82.04 & 70.88 & 80.41 & 56.04 \\
32 chunks per height & 81.25 & 70.90 & 82.45 & 56.20 \\
128 chunks per height & 80.36 & 70.70 & \textbf{83.20} & 55.05 \\
\specialrule{1.2pt}{0pt}{0pt}
\end{tabular}%
}
\caption{Ablation results on tree encoder and memory construction. All results are on Qwen-2.5-7B with temperature=0 and do\_sample=False.}
\label{tab:appendixAdditional}
\end{table}

\section{Model Size and Compute}
\label{appendix1}
\label{sec:appendix}
\subsection{Model Variants and Parameter Counts}
For computational efficiency, the reported results for HellaSwag and WinoGrande are evaluated on the validation split. And Appendix Table \ref{tab:appendix1} reports the parameter counts and key configuration differences for each variant. 

\begin{table}[t]
    \centering
    \resizebox{\columnwidth}{!}{
    \begin{tabular}{lll}
    \specialrule{1.2pt}{0pt}{0pt}
        Variant  & Backbone  & Trainable Params  \\ \hline
        Backbone & Qwen-2.5-7B & /  \\ 
        LoRA-only & Qwen-2.5-7B & 0.01B  \\ 
        Direct-Joint & Qwen-2.5-7B & 1.55B  \\ 
        GTCA & Qwen-2.5-7B & 1.55B  \\ 
        Backbone & LLaMA-3-8B & /  \\ 
        LoRA-only & LLaMA-3-8B & 0.01B  \\ 
        Direct-Joint & LLaMA-3-8B & 2.43B   \\ 
        GTCA & LLaMA-3-8B & 2.43B  \\ 
        \specialrule{1.2pt}{0pt}{0pt}
    \end{tabular}
    }
      \caption{Model variants and sizes.}
  \label{tab:appendix1}
\end{table}

\subsection{Overall Compute Budget and Computing Infrastructure}
All continued training and evaluation runs were conducted with a fixed compute budget. Each run was trained with sequence length 1024. Unless otherwise noted, we reported mean results over five runs with 5 different random seeds.

Training was implemented in PyTorch using DistributedDataParallel (DDP) across 8 GPUs per run. The total compute budget was approximately 200–300 GPU-hours. We used bf16 mixed precision and the AdamW optimizer. Relevant software versions are as follows: CUDA 12.8, cuDNN 9.10.2, PyTorch 2.8, NCCL 2.27.3, nltk 3.9.1, and spacy 3.8.7.

\section{Prompts Used for Benchmark}
\label{appendixB}
We used only publicly available benchmark datasets and pretrained checkpoints and followed their original licenses and terms of use. The pretrained backbones are used under Apache License 2.0 for Qwen2.5 and the Meta LLaMA 3 Community License for LLaMA 3, and the benchmark datasets are used under their respective licenses such as MIT, Apache License or Creative Commons, with CLOTH restricted to non-commercial research use. We did not redistribute any dataset content, or model weights, and any derived models are intended solely as research prototypes for controlled evaluation consistent with the access conditions of the underlying resources.

Below are the prompts for each benchmark. For k-shot settings, k demonstration examples were prepended verbatim before the test instance.

\subsection{CLOTH (0-shot) and MMLU (0-shot)} 

Instruction: Choose the correct option based on the question. Output the full text of the chosen option exactly as it appears under Options.\\
Question: \{Question\}\\
Options: \\
A.\ \{option\_A\} \\
B.\ \{option\_B\} \\
C.\ \{option\_C\} \\
D.\ \{option\_D\}\\
Answer:
\subsection{BLiMP (minimal pairs, pairwise preference)} 
Prompt template (no answer token):\\
\{sentence\_good\}\\
\{sentence\_bad\}\\
This evaluation uses two separate forward passes, scoring \{sentence\_good\}  and \{sentence\_bad\} independently.
\subsection{CoLA (0-shot, binary classification as MCQA)}
Instruction: You are a linguist. Decide if the following English sentence is grammatically acceptable. Output 1 for acceptable, 0 for unacceptable. Output only a single character: 0 or 1.\\
Sentence: \{sentence\}\\
Answer:
\subsection{HellaSwag (10-shot)}
Instruction: Choose the correct option based on the question. Output the full text of the chosen option exactly as it appears under Options.\\
Demonstration format (repeated 10 times)\\
Context: Read the context and choose the most plausible continuation. \\
\{context\}\\
Options:\\
A. \{ending0\}\\
B. \{ending1\}\\
C. \{ending2\}\\
D. \{ending3\}\\
Answer: \{gold\_answer\}\\
Context: Read the context and choose the most plausible continuation. \\
\{context\}\\
Options:\\
A. \{ending0\}\\
B. \{ending1\}\\
C. \{ending2\}\\
D. \{ending3\}\\
Answer: 
\subsection{WinoGrande (5-shot)}
Instruction: Choose the correct option based on the question. Output the full text of the chosen option exactly as it appears under Options.\\
Demonstration format (repeated 5 times)\\
Question: \{Question\}\\
Options: \\
A.\ \{option\_A\} \\
B.\ \{option\_B\}\\
Answer: \{gold\_answer\}\\
Question: \{Question\}\\
Options: \\
A.\ \{option\_A\} \\
B.\ \{option\_B\}\\
Answer:

\section{\texorpdfstring{Grid Search for the Structural Coefficient $\alpha_{\text{struct}}$}{Grid Search for the structural coefficient alpha\textunderscore struct}}
\label{appendixD}

GTCA injects parse tree information through a gated tree cross-attention branch and integrates the resulting structural update via a token-update-masked residual weighted by a structural coefficient $\alpha_{\text{struct}}$. In practice, $\alpha_{\text{struct}}$ controls the strength of structure injection and therefore mediates a trade-off between (i) syntactic specialization and (ii) retention of general-purpose capability. We select the target $\alpha_{\text{struct}}$ via grid search, as stated in the main experiments section.

We grid search over a discrete set of target values, a recommended starting grid is:
$\alpha_{\text{struct}}^* \in \{0.00, 0.05, 0.10, 0.15, 0.20\}$.
For each candidate $\alpha_{\text{struct}}^*$, we use the same staged training pipeline as Appendix Table \ref{tab:aa} in Section\ref{Sec 3.4}, with:
\begin{itemize}
    \item Stage 1: $\alpha_{\text{struct}} = 0$ (structure path disabled).
    \item Stage 2: linearly warm up $\alpha$ from $0$ to $\alpha_{\text{struct}}^*$ over the warm-up steps.
    \item Stage 3: keep $\alpha$ fixed at $\alpha_{\text{struct}}^*$.
\end{itemize}
An explicit form of the warm-up can be written as:
\[
\alpha(t) = \min\left( \alpha_{\text{struct}}^*,\alpha_{\text{struct}}^* \cdot \frac{t}{T_{\text{warm}}} \right)
\]
where $t$ is the current optimization step within Stage 2 and $T_{\text{warm}}$ is the warm-up length.

We select $\alpha_{\text{struct}}^*$ using held-out validation performance under a retention-aware criterion. Appendix Table \ref{tab:aa} reports the validation results across candidate $\alpha_{\text{struct}}$ values.

\begin{table}[t]
\centering
\resizebox{\columnwidth}{!}{%
\begin{tabular}{lcccc}
\specialrule{1.2pt}{0pt}{0pt}
\multirow{2}{*}{$\alpha_{\text{struct}}$} &
\multicolumn{2}{c}{Multiple-Choice QA} &
\multicolumn{2}{c}{Syntactic Robustness} \\
\cmidrule(lr){2-3}\cmidrule(lr){4-5}
& \begin{tabular}[c]{@{}c@{}}CLOTH\\(0-shot)\end{tabular}
& \begin{tabular}[c]{@{}c@{}}MMLU\\(0-shot)\end{tabular}
& BLiMP
& \begin{tabular}[c]{@{}c@{}}CoLA\\(0-shot)\end{tabular} \\
\specialrule{0.8pt}{0pt}{0pt}
0.05 & 81.35          & 67.78          & 74.65          & 53.65 \\
0.10 & \textbf{82.51} & 67.12          & 75.32          & 52.14 \\
0.15 & 81.46          & \textbf{69.36} & \textbf{79.63} & \textbf{55.20} \\
0.20 & 80.36          & 65.35          & 80.12          & 55.14 \\
\specialrule{1.2pt}{0pt}{0pt}
\end{tabular}%
}
\caption{Grid search over $\alpha$ on the validation split.  All results are on Qwen-2.5-7B, and we set temperature=0 and disabled sampling (do\_sample=False).}
\label{tab:aa}
\end{table}

Across the grid, small $\alpha_{\text{struct}}$ values tend to under-utilize structural memory, while overly large $\alpha_{\text{struct}}$ can introduce interference that degrades retention. The selected $\alpha_{\text{struct}}$ operates in a regime where structure injection is strong enough to improve targeted syntactic judgments yet remains compatible with preserving general-purpose capability. We finally select $\alpha_{\text{struct}}^* = 0.15$, because it achieves the best trade-off between syntactic gains (BLiMP) and retention (MMLU) under the criterion above.

\section{Error Analysis}
\label{appendixerror}

Across the BLiMP phenomena that directly probe word order and movement, the residual errors concentrate in filler-gap dependencies and island effects. These cases are informative because the lexical content is largely preserved while the syntactic configuration changes. The examples below show that the model can still over-rely on local surface compatibility and can underweight global constraints that require tracking long-distance dependencies or respecting extraction islands. This points to clear directions for improvement, such as training objectives that better enforce global constraint satisfaction and evaluation protocols that stress paired stability under controlled syntactic transformations.

\paragraph{Examples of residual errors on filler-gap dependencies and island effects.}
For each phenomenon, we show a grammatical sentence, an ungrammatical sentence with a minimal structural change, and the model's preference. The model consistently prefers the ungrammatical version, indicating overreliance on local surface compatibility and underweighting of global syntactic constraints.

\begin{enumerate}
    \item \textbf{Filler-gap dependency}
    
    \textit{Grammatical:} Kendra saw some ice cream that Maria isn't hiding.
    
    \textit{Ungrammatical:} Kendra saw what Maria isn't hiding some ice cream.
    
    \textit{Model preference:} prefers ungrammatical.

    \item \textbf{Filler-gap dependency}
    
    \textit{Grammatical:} Gina had known that person that Dan hadn't forgotten that had criticized many dancers.
    
    \textit{Ungrammatical:} Gina had known who that person that Dan hadn't forgotten had criticized many dancers.
    
    \textit{Model preference:} prefers ungrammatical.

    \item \textbf{Filler-gap dependency}
    
    \textit{Grammatical:} Timothy can conceal who those guys that come here weren't scaring.
    
    \textit{Ungrammatical:} Timothy can conceal that those guys that come here weren't scaring.
    
    \textit{Model preference:} prefers ungrammatical.

    \item \textbf{Island effects}
    
    \textit{Grammatical:} Who wouldn't Mitchell investigate he is kissing?
    
    \textit{Ungrammatical:} Who wouldn't Mitchell investigate who is kissing?
    
    \textit{Model preference:} prefers ungrammatical.

    \item \textbf{Island effects}
    
    \textit{Grammatical:} Who would actresses' hiding most guests stun.
    
    \textit{Ungrammatical:} Who would actresses' hiding stun most guests.
    
    \textit{Model preference:} prefers ungrammatical.
\end{enumerate}

\begin{table}[t]
\centering
\resizebox{\columnwidth}{!}{%
\begin{tabular}{lccc}
\specialrule{1.2pt}{0pt}{0pt}
\multirow{2}{*}{Model} &
\multicolumn{3}{c}{Controlled Grammatical Perturbation on CLOTH} \\
\cmidrule(lr){2-4}
& \begin{tabular}[c]{@{}c@{}}Original\\accuracy\end{tabular}
& \begin{tabular}[c]{@{}c@{}}Perturbed\\accuracy\end{tabular}
& Consistency \\
\specialrule{0.8pt}{0pt}{0pt}
Backbone     & 76.21          & 72.34          & 91.90          \\
LoRA-only    & 78.47          & 74.76          & 91.05          \\
Direct-Joint & 80.11          & 79.02          & 94.18          \\
GTCA         & \textbf{81.68} & \textbf{80.42} & \textbf{98.27} \\
\specialrule{1.2pt}{0pt}{0pt}
\end{tabular}%
}
\caption{Performance under controlled grammatical perturbations on 1000 randomly sampled CLOTH questions. Perturbations preserve the gold answer while altering surface syntax. Consistency denotes the fraction of items whose predicted option remains unchanged across the original and perturbed questions. All results are on Qwen-2.5-7B, and we set temperature=0 and disabled sampling (do\_sample=False).}
\label{tab:appendix_perturb}
\end{table}

\begin{table}[t]
\centering
\resizebox{\columnwidth}{!}{%
\begin{tabular}{lccc}
\specialrule{1.2pt}{0pt}{0pt}
\multirow{2}{*}{Model} &
\multicolumn{3}{c}{Controlled Surface Noise on MMLU} \\
\cmidrule(lr){2-4}
& \begin{tabular}[c]{@{}c@{}}Clean\\accuracy\end{tabular}
& \begin{tabular}[c]{@{}c@{}}Noisy\\accuracy\end{tabular}
& Consistency \\
\specialrule{0.8pt}{0pt}{0pt}
Backbone     & 67.50          & 64.80          & 82.67          \\
LoRA-only    & 69.10          & 66.00          & 82.60          \\
Direct-Joint & 70.20          & 68.60          & 84.43          \\
GTCA         & \textbf{72.00} & \textbf{71.10} & \textbf{88.95} \\
\specialrule{1.2pt}{0pt}{0pt}
\end{tabular}%
}
\caption{Performance under controlled surface noise on 1000 randomly sampled MMLU questions. Noise is applied only to the question region and preserves the intended meaning. Consistency denotes the fraction of items whose predicted option remains unchanged across the clean and noisy questions. All results are on Qwen-2.5-7B, and we set temperature=0 and disabled sampling (do\_sample=False).}
\label{tab:appendix_noise}
\end{table}

\section{Robustness under Controlled Perturbations and Surface Noise}
\label{appendixtree}
We construct a controlled grammatical perturbation suite on 1000 randomly sampled CLOTH questions. We apply meaning preserving transformations to the question region, including active passive alternation, relative clause insertion, and adjunct reordering, while keeping the gold answer unchanged. As shown in Appendix Table \ref{tab:appendix_perturb}, GTCA achieves the highest perturbed accuracy and the highest consistency. This indicates that GTCA reduces answer flips under syntactic variation and better preserves decision stability when surface form changes but meaning remains constant.

We further test robustness to meaning preserving surface noise, including punctuation deletion, mild typos, and whitespace corruption, on 1000 randomly sampled MMLU questions. Appendix Table \ref{tab:appendix_noise} shows that GTCA also yields the highest noisy accuracy and the highest consistency under this setting. These results suggest that the structural pathway remains beneficial even when parser inputs are mildly degraded, although the method still relies on parser quality as discussed in the limitations.

\end{document}